\documentclass{article} % For LaTeX2e
\usepackage{iclr2027_conference,times}

\usepackage{amsmath,amsfonts,bm}

\def\eqref#1{equation~\ref{#1}}
\def\1{\bm{1}}

\DeclareMathAlphabet{\mathsfit}{\encodingdefault}{\sfdefault}{m}{sl}
\SetMathAlphabet{\mathsfit}{bold}{\encodingdefault}{\sfdefault}{bx}{n}

\usepackage{hyperref}
\usepackage{url}
\usepackage{graphicx}
\usepackage{booktabs}
\usepackage{longtable}
\usepackage{verbatim}
\usepackage{amsmath}
\usepackage{microtype}
\usepackage{tikz}
\usetikzlibrary{arrows.meta,positioning}
\definecolor{pnavy}{RGB}{43,74,120}
\definecolor{pgray}{RGB}{106,111,116}
\definecolor{pcarm}{RGB}{164,36,59}
\definecolor{pfill}{RGB}{244,246,249}
\hypersetup{colorlinks=true, linkcolor=pnavy, citecolor=pnavy, urlcolor=pnavy}

\title{Measuring the Checker: Mutation Analysis \\ for GPU-Kernel Benchmark Oracles}

\author{Mingzhe Du$^{1,2}$ \quad Anh Tuan Luu$^{2,3}$ \quad Dong Huang$^{1}$ \quad See-Kiong Ng$^{1}$ \\[3pt]
$^{1}$National University of Singapore \hspace{0.15em}
$^{2}$Nanyang Technological University \hspace{0.15em}
$^{3}$CAIR, VinUniversity \\[2pt]
\texttt{\string{mingzhe,\,dhuang,\,seekiong\string}@nus.edu.sg}, \texttt{anhtuan.luu@ntu.edu.sg}
}

\iclrfinalcopy
\begin{document}

\maketitle
\lhead{Preprint}

\begin{abstract}
Benchmarks for LLM-generated GPU kernels decide correctness with a few random
inputs and a loose floating-point tolerance, and their verdicts now
feed leaderboards and reinforcement-learning rewards. Recent work agrees these
checkers are weak and patches them by hand---extra input distributions,
fuzzing recipes, tighter tolerances---with no way to \emph{measure} whether
any patch suffices. We introduce mutation analysis as an adequacy metric for
kernel-benchmark oracles: deterministic rules inject 10{,}303 compilable
faults into verified CUDA implementations of 188 KernelBench problems,
7{,}384 of them with an independent kill witness; any test protocol is
scored by the fraction it detects. The official check misses \textbf{one
in six} witnessed faults (16.9\%), deterministically, and the misses are skewed by family: 8.7\% of
arithmetic faults escape but 78.6\% of precision faults do. The metric explains why (a tolerance blind band growing with reduction size; a measured ceiling on input aggressiveness set by legitimate
floating-point variance), audits the strongest existing patch
(KernelBench-Verified's gain splits into $+4.0$ points from hidden inputs and
$+4.5$ from tighter tolerance, a split its authors could not compute), and exposes a published fuzzing recipe that rejects \emph{correct}
kernels 107 times. Optimizing suites over the kill matrix reaches 98.0\%
detection with two inputs per problem (94.8\% held-out), and the
measurement's fault taxonomy teaches a test generator more than the raw
faults themselves. Across 48 whole architectures the blindness grows with
scale, concentrating in deep homogeneous
pipelines, and two problems prove unrefereeable: their official references
violate the benchmark's own tolerance against fp64. We release everything as
\href{https://huggingface.co/datasets/Elfsong/KernelBench-M}{KernelBench-M}.
\end{abstract}

\section{Introduction}
\label{sec:intro}

A benchmark's correctness checker used to be bookkeeping. For GPU-kernel
generation it has become infrastructure that carries load: KernelBench-style
verdicts \citep{ouyang2025kernelbench} rank models publicly, gate which
generated kernels reach production experiments, and---most consequentially---
serve as the reward signal for reinforcement learning systems that write
kernels \citep{baronio2025kevin}. A weak checker in this position fails
silently. The model does not learn to write correct kernels; it learns to
write kernels that \emph{pass}, and the two diverge exactly where the checker
is blind.

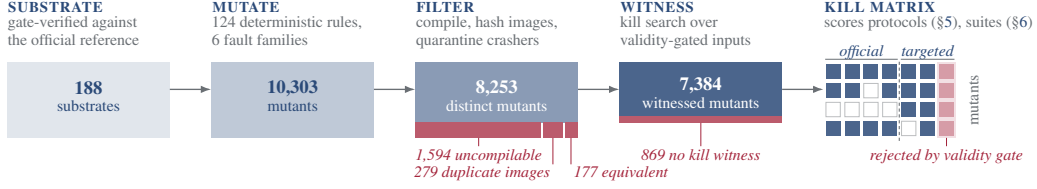
\begin{figure}[t]
\centering
\begin{tikzpicture}[x=1cm,y=1cm,
  hdr/.style={font=\scriptsize\bfseries, text=pnavy, anchor=south west, inner sep=0pt},
  sub/.style={font=\tiny, text=pgray, anchor=north west, inner sep=0pt, align=left},
  num/.style={font=\scriptsize\bfseries, inner sep=0pt},
  lab/.style={font=\tiny, inner sep=0pt},
  rem/.style={font=\tiny, text=pcarm, inner sep=0pt, align=left, anchor=north west},
  arr/.style={-{Latex[length=1.4mm,width=1.0mm]}, line width=0.5pt, draw=pgray},
  rarr/.style={-{Latex[length=1.2mm,width=0.9mm]}, line width=0.5pt, draw=pcarm}]
% ---- geometry: blocks top-aligned; height proportional to mutant count ----
\def\yt{1.55}
\def\hA{1.00}\def\hB{0.801}\def\hC{0.717}          % 10,303 : 8,253 : 7,384
\pgfmathsetmacro{\ybA}{\yt-\hA}\pgfmathsetmacro{\ybB}{\yt-\hB}\pgfmathsetmacro{\ybC}{\yt-\hC}
\pgfmathsetmacro{\ya}{\yt-\hC/2}                    % common arrow height
\def\bw{2.15}\def\gap{0.55}
\def\xS{0.0}\pgfmathsetmacro{\xSe}{\xS+\bw}
\pgfmathsetmacro{\xM}{\xSe+\gap}\pgfmathsetmacro{\xMe}{\xM+\bw}
\pgfmathsetmacro{\xF}{\xMe+\gap}\pgfmathsetmacro{\xFe}{\xF+\bw}
\pgfmathsetmacro{\xW}{\xFe+\gap}\pgfmathsetmacro{\xWe}{\xW+\bw}
\pgfmathsetmacro{\xK}{\xWe+\gap}
\def\hy{2.21}   % header baseline
% ---- stage blocks ----
\fill[pnavy!14] (\xS,\ybA) rectangle (\xSe,\yt);
\fill[pnavy!30] (\xM,\ybA) rectangle (\xMe,\yt);
\fill[pnavy!55] (\xF,\ybB) rectangle (\xFe,\yt);
\fill[pnavy!85] (\xW,\ybC) rectangle (\xWe,\yt);
\node[num, text=pnavy] at ({(\xS+\xSe)/2},{\yt-0.34}) {188};
\node[lab, text=pnavy] at ({(\xS+\xSe)/2},{\yt-0.58}) {substrates};
\node[num, text=pnavy] at ({(\xM+\xMe)/2},{\yt-0.34}) {10{,}303};
\node[lab, text=pnavy] at ({(\xM+\xMe)/2},{\yt-0.58}) {mutants};
\node[num, text=white] at ({(\xF+\xFe)/2},{\yt-0.34}) {8{,}253};
\node[lab, text=white] at ({(\xF+\xFe)/2},{\yt-0.58}) {distinct mutants};
\node[num, text=white] at ({(\xW+\xWe)/2},{\yt-0.30}) {7{,}384};
\node[lab, text=white] at ({(\xW+\xWe)/2},{\yt-0.53}) {witnessed mutants};
% ---- flow arrows (one common height) ----
\draw[arr] (\xSe,\ya) -- (\xM,\ya);
\draw[arr] (\xMe,\ya) -- (\xF,\ya);
\draw[arr] (\xFe,\ya) -- (\xW,\ya);
\draw[arr] (\xWe,\ya) -- (\xK,\ya);
% ---- removed slices: the strip cut off at each step sits flush beneath the
%      surviving block (same width, height = height lost), split by cause ----
% FILTER: 10,303 -> 8,253 ; pieces 1,594 / 279 / 177 (widths proportional)
\pgfmathsetmacro{\stF}{\ybB}\pgfmathsetmacro{\sbF}{\stF-(\hA-\hB)}
\pgfmathsetmacro{\wA}{\bw*1594/2050}\pgfmathsetmacro{\wB}{\bw*279/2050}\pgfmathsetmacro{\wC}{\bw*177/2050}
\fill[pcarm!75] (\xF,\sbF) rectangle ({\xF+\wA-0.015},\stF);
\fill[pcarm!75] ({\xF+\wA+0.015},\sbF) rectangle ({\xF+\wA+\wB-0.015},\stF);
\fill[pcarm!75] ({\xF+\wA+\wB+0.015},\sbF) rectangle (\xFe,\stF);
% WITNESS: 8,253 -> 7,384 ; one piece, 869
\pgfmathsetmacro{\stW}{\ybC}\pgfmathsetmacro{\sbW}{\stW-(\hB-\hC)}
\fill[pcarm!75] (\xW,\sbW) rectangle (\xWe,\stW);
% Labels: wide pieces get a centred label under a short tick; the two narrow
% pieces get callouts to the lower right (levels chosen so no leader crosses text).
\def\lvA{0.41}\def\lvB{0.17}\def\lvC{-0.07}   % label tops
\tikzset{lead/.style={draw=pcarm, line width=0.35pt},
         rlab/.style={font=\tiny\itshape, text=pcarm, inner sep=0pt}}
\pgfmathsetmacro{\pa}{\xF+\wA/2}\pgfmathsetmacro{\pb}{\xF+\wA+\wB/2}\pgfmathsetmacro{\pc}{\xF+\wA+\wB+\wC/2}
\pgfmathsetmacro{\pd}{(\xW+\xWe)/2}
\draw[lead] (\pa,\sbF) -- (\pa,\lvA+0.03); \node[rlab, anchor=north] at (\pa,\lvA) {1{,}594 uncompilable};
\draw[lead] (\pd,\sbW) -- (\pd,\lvA+0.03); \node[rlab, anchor=north] at (\pd,\lvA) {869 no kill witness};
\draw[lead] (\pb,\sbF) -- (\pb,\lvB+0.03); \node[rlab, anchor=north east] at (\pb-0.03,\lvB+0.02) {279 duplicate images};
\draw[lead] (\pc,\sbF) -- (\pc,\lvB+0.03); \node[rlab, anchor=north west] at (\pc+0.03,\lvB+0.02) {177 equivalent};
% ---- headers, all left-aligned on their block ----
\node[hdr] at (\xS,\hy) {\textsc{substrate}};
\node[sub] at (\xS,{\hy-0.04}) {gate-verified against\\the official reference};
\node[hdr] at (\xM,\hy) {\textsc{mutate}};
\node[sub] at (\xM,{\hy-0.04}) {124 deterministic rules,\\6 fault families};
\node[hdr] at (\xF,\hy) {\textsc{filter}};
\node[sub] at (\xF,{\hy-0.04}) {compile, hash images,\\quarantine crashers};
\node[hdr] at (\xW,\hy) {\textsc{witness}};
\node[sub] at (\xW,{\hy-0.04}) {kill search over\\validity-gated inputs};
\node[hdr] at (\xK,\hy) {\textsc{kill matrix}};
\node[sub] at (\xK,{\hy-0.04}) {scores protocols (\S\ref{sec:audit}), suites (\S\ref{sec:synthesis})};
% ---- kill matrix: top-aligned with the blocks, as tall as the tallest ----
\def\cw{0.25}\def\nr{4}\def\nc{7}
\pgfmathsetmacro{\yM}{\yt-\nr*\cw}
\pgfmathsetmacro{\xKe}{\xK+\nc*\cw}\pgfmathsetmacro{\yMe}{\yM+\nr*\cw}
\pgfmathsetmacro{\cxl}{\xK+(\nc-1)*\cw}
\fill[pcarm!14] (\cxl,\yM) rectangle (\xKe,\yMe);
\foreach \r/\row in {0/{1,1,1,1,1,1,1},1/{1,1,0,1,1,1,1},2/{0,0,0,0,1,1,1},3/{1,1,1,1,0,1,1}}{
  \foreach \v [count=\k from 0] in \row {
    \pgfmathsetmacro{\cx}{\xK+\k*\cw}\pgfmathsetmacro{\cy}{\yMe-(\r+1)*\cw}
    \ifnum\k=6
      \ifnum\v=1 \fill[pcarm!45] (\cx+0.03,\cy+0.03) rectangle (\cx+\cw-0.03,\cy+\cw-0.03);
      \else \draw[pcarm!45, line width=0.3pt] (\cx+0.03,\cy+0.03) rectangle (\cx+\cw-0.03,\cy+\cw-0.03); \fi
    \else
      \ifnum\v=1 \fill[pnavy!85] (\cx+0.03,\cy+0.03) rectangle (\cx+\cw-0.03,\cy+\cw-0.03);
      \else \draw[pgray!45, line width=0.3pt] (\cx+0.03,\cy+0.03) rectangle (\cx+\cw-0.03,\cy+\cw-0.03); \fi
    \fi
  }}
\draw[pgray, line width=0.35pt] (\xK+0.02,{\yMe+0.04}) -- ({\xK+4*\cw-0.02},{\yMe+0.04});
\node[font=\tiny\itshape, text=pnavy, anchor=base] at ({\xK+2*\cw},{\yMe+0.085}) {official};
\draw[pgray, line width=0.35pt] ({\xK+4*\cw+0.02},{\yMe+0.04}) -- ({\xK+7*\cw-0.02},{\yMe+0.04});
\node[font=\tiny\itshape, text=pnavy, anchor=base] at ({\xK+5.5*\cw},{\yMe+0.085}) {targeted};
\node[font=\tiny, text=pgray, anchor=north, rotate=90] at (\xKe+0.05,{(\yM+\yMe)/2}) {mutants};
% dashed separator between official and targeted inputs
\draw[pgray, line width=0.4pt, dash pattern=on 1.2pt off 1pt] ({\xK+4*\cw},{\yM-0.04}) -- ({\xK+4*\cw},{\yMe+0.04});
\pgfmathsetmacro{\pe}{\cxl+\cw/2}
\draw[lead] (\pe,\yM) -- (\pe,\lvA+0.03); \node[rlab, anchor=north] at (\pe,\lvA) {rejected by validity gate};
\end{tikzpicture}
\caption{The measurement pipeline. Block height is proportional to mutant
count. Deterministic rules mutate 188 gate-verified CUDA substrates into
10{,}303 mutants; compilation, compiled-image hashing, and crash quarantine
remove 2{,}050 (the red strip cut from beneath each block, split by cause), and a kill-witness search over validity-gated
inputs admits 7{,}384 into the scoring denominator, quarantining the 869 that
no valid input kills. The witnessed kill matrix (mutants $\times$ inputs,
filled = detected) records only inputs that pass the validity gate; an input
that rejects the correct kernel (red column) is discarded. The matrix then
scores any protocol, synthesizes minimal suites, and yields the fault
taxonomy.}
\label{fig:pipeline}
\end{figure}

The community has noticed. KernelBench-Verified \citep{zhang2026kbverified}
documents generated kernels that hard-code their way past the official check,
adds four hidden input distributions and a tighter tolerance, and reports that
headline speedups collapse from $1.43\times$ to $0.88\times$ under the
stronger protocol. The Correctness Illusion \citep{sarkar2026illusion} reaches
the same verdict by seeding nine bugs by hand and proposing a fuzzing oracle;
robust-kbench \citep{lange2025robust} hardens input shapes and timing
methodology. Each of these efforts \emph{patches} the checker. None of them
can answer the question a benchmark maintainer actually faces: \emph{does my
patch cover the faults that matter, and what does it still miss?} Validating a
protocol against ten hand-seeded bugs measures little; as we show in
\S\ref{sec:audit}, a patch can look decisive on anecdotes while entire fault
families remain unreachable \emph{by construction}.

Software engineering solved this measurement problem fifty years ago.
Mutation analysis \citep{demillo1978hints,jia2011survey} scores a test suite
by the fraction of small injected faults it detects, turning ``is my test
suite good?'' from opinion into measurement. But the standard recipe does not
transfer directly to GPU kernels: the oracle is a graded numerical comparison
rather than an exact one, aggressive test inputs can reject \emph{correct}
implementations through legitimate floating-point variance, equivalent and
oracle-blind mutants pollute the denominator, and per-mutant compilation cost
makes naive matrices intractable (\S\ref{sec:method}). Our central
contribution is a mutation-analysis methodology that survives contact with
these obstacles (Fig.~\ref{fig:pipeline})---and what the resulting metric
reveals about the benchmarks the field is standing on.

Concretely:
\textbf{(i)}~We build the first at-scale measurement of kernel-benchmark
oracle strength: 10{,}303 rule-generated faults across 188 verified CUDA
substrates, 7{,}384 of them backed by kill witnesses so that only
provably-detectable mutants enter any denominator (prior art validates
against $\le 10$ hand-seeded bugs). The official KernelBench check misses
16.9\% of witnessed faults---one in six---stably across scale and
deterministically, with misses concentrated by family: 78.6\% of precision
faults and 27.8\% of synchronization faults escape, against 8.7\% of textbook
arithmetic mutations (Table~\ref{tab:tax}).
\textbf{(ii)}~We identify and quantify the domain's two governing mechanisms:
\emph{tolerance vacuity}, a blind band that grows with reduction size until an
all-zeros output passes the check, and the \emph{legitimate-variance ceiling},
a measured bound above which harsher inputs reject correct kernels
(Fig.~\ref{fig:mechanisms}).
\textbf{(iii)}~We audit existing patches under their own published parameters
(\S\ref{sec:audit}): KernelBench-Verified's improvement decomposes into
$+4.0$ points from its four hidden distributions and $+4.5$ from its tighter
tolerance---a decomposition its authors state they cannot compute---while its
constant-scaling design leaves shape- and indexing-fault families unreachable;
a reconstructed fuzzing baseline crosses the variance ceiling and falsely
rejects correct kernels 107 times.
\textbf{(iv)}~We show measurement enables synthesis (\S\ref{sec:synthesis}):
set cover over the kill matrix yields two-input suites at 98.0\% detection
(94.8\% held-out) against 83.1\% for the official five; and a
knowledge-ladder experiment shows that giving a test generator the
measurement-derived fault \emph{taxonomy} nearly triples a fuzz baseline and
outperforms giving it the concrete faults themselves.
\textbf{(v)}~We extend substrates to whole architectures (\S\ref{sec:depth}):
across 48 networks and 12{,}361 distinct mutants, misses exceed operator
level on every accounting (strict: 17.3\% vs.\ 16.9\%; upper bound:
$1.7\times$), survival concentrates in deep homogeneous pipelines while
normalization-dense networks stay checkable, and two further problems prove
\emph{unrefereeable}---their official references violate the benchmark's own
tolerance against their fp64 selves. As generated kernels grow from operators
into models, the checking problem gets harder, not easier.

\section{Two faults the official check cannot see}
\label{sec:motivation}

% Neither example requires trusting our framework.

\textbf{A zeros-output softmax passes.} One KernelBench problem applies
softmax across $d = 393{,}216$ elements. Under the official
\texttt{torch.rand} inputs the reference output averages $2.5\times10^{-6}$
per element, while the check accepts any output within
$\texttt{atol}=10^{-2}$---four thousand times the signal. A kernel that
returns all zeros passes every official trial. The blind band is not an edge
case; it widens systematically with reduction size
(Fig.~\ref{fig:mechanisms}b).

\textbf{No reseeding can help.} The official inputs are drawn from $[0,1)$.
Every element is positive, so deleting a ReLU is the identity on the entire
support; $\exp(x)$ cannot overflow for $x<1$, so removing softmax's
max-subtraction stabilizer is unobservable. Survival under such inputs is a
property of the \emph{distribution}, not of sampling luck---more random trials
measure the same blindness more confidently.

\section{Mutation analysis for graded numerical oracles}
\label{sec:method}

\paragraph{Setting.} A benchmark problem supplies a PyTorch reference $f$ and
an input generator; a submission $\hat f$ passes if
$\texttt{allclose}(f(x), \hat f(x); \texttt{atol}, \texttt{rtol})$ on a few
draws. We want to score the \emph{protocol}---inputs plus tolerance---by the
fraction of faults it detects. Four obstacles separate this domain from
classical mutation testing and from test-augmentation work on Python
benchmarks \citep{liu2023evalplus}.

\begin{figure}[t]
\centering
\includegraphics[width=\linewidth]{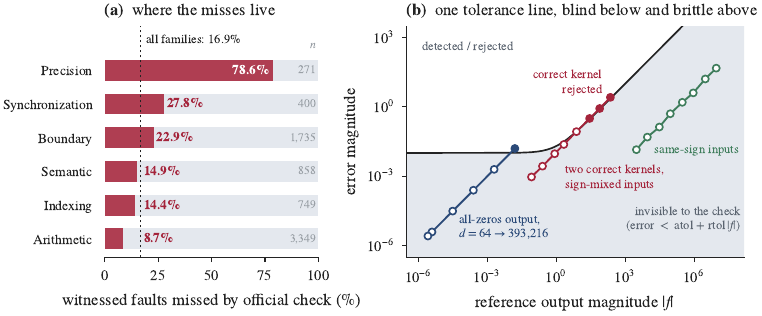}
\caption{What the official check misses, and why. \textbf{(a)}~Share of
witnessed faults missed by the official protocol, by family ($n$ at right):
textbook arithmetic mutations are mostly caught; precision faults escape at
$9\times$ the average rate. \textbf{(b)}~The check accepts any error below
$\texttt{atol} + \texttt{rtol}\,|f|$ (black line; grey region invisible to
it; filled markers: the check fires, open: it accepts). \emph{Vacuity:} the fault ``return all zeros'' has error $|f|$ (blue);
as the softmax reduction $d$ grows it slides into the grey region and passes.
\emph{Validity ceiling:} the fp32 accumulation noise between two
\emph{correct} kernels grows with input scale $s$; on sign-mixed inputs the
output cancels and the noise climbs out of the region, rejecting a correct
kernel ($s \ge 300$); on same-sign inputs it stays deep inside
(green).}
\label{fig:mechanisms}
\end{figure}

\textbf{The oracle is graded (C1).} Detection depends not on behavioural
difference but on the ratio of a fault's error to the \emph{output magnitude}
entering the tolerance. We measure the consequence: at $d=64$ the largest
undetectable relative error is $0.39$; at $d=393{,}216$ it is
$2.5\times10^{3}$ (Fig.~\ref{fig:mechanisms}b).

\textbf{Valid inputs are bounded above (C2).} Two \emph{correct} fp32 kernels
legitimately disagree through accumulation order
\citep{goldberg1991floating,shanmugavelu2024fpna}. On a $K{=}4096$ reduction,
sign-mixed inputs scaled by $300$ push a correct kernel past the official
tolerance---the input itself becomes invalid---while same-sign inputs of far
larger magnitude remain safe, because the tolerance scales with the
un-cancelled output (Fig.~\ref{fig:mechanisms}b). Every input-generation
scheme for this domain needs a validity gate located by this ceiling; we found
none in prior work that has one, and \S\ref{sec:audit} shows a published
recipe paying the price.

\textbf{The denominator must be earned (C3).} Some mutants are equivalent (a
coherent \texttt{blockIdx} permutation relabels independent work); some are
real faults no numerical oracle can see (an out-of-bounds write landing in
allocator slack). Scoring suites against unkillable rows deflates every
protocol equally and informs about none. We therefore admit a mutant into the
denominator only with a \emph{kill witness}---a concrete, validity-gated input
on which it verifiably fails---and quarantine the rest. Where we additionally
report survival over all behaviourally \emph{distinct} mutants (which includes
the not-yet-adjudicated), we label it explicitly; that looser rate upper-bounds
oracle blindness.

\textbf{The Cost (C4).} Compiling one mutant through the standard
\texttt{torch} extension path takes ${\sim}200$\,s; ten thousand mutants would
need GPU-years. Runtime compilation~(NVRTC) brings this to 84\,ms---a $500\times$ reduction that
makes the kill matrix affordable.

\paragraph{Pipeline.} Fig.~\ref{fig:pipeline} summarizes the system.
\emph{Substrates:} mutation needs source, and KernelBench's references bottom
out in closed cuDNN/cuBLAS binaries; for each problem we maintain a correct
CUDA implementation used \emph{only} as a mutation target---the oracle remains
the benchmark's own reference. Substrates are LLM-authored and admitted by an
automated gate that checks them against the official reference on every suite
(the gate rejects real errors, e.g.\ an L1-norm dividing by sum instead of
mean). \emph{Mutants:} 124 deterministic rules spanning six families
(Table~\ref{tab:tax}): classical arithmetic/relational replacements;
GPU-specific operators in the lineage of \citet{zhu2020mutgpu} and
\citet{bradbury2006concurrent}---barrier removal,
\texttt{\_\_syncthreads}${\to}$\texttt{\_\_syncwarp}, ceil-to-floor grid
division, bounds-guard deletion, fp16 accumulation, index-axis swaps---and
LLM-mined fine-grained families (argmax tie-breaking, perturbed polynomial
constants, shifted piecewise thresholds). \emph{Filters:} NVRTC rejects
1{,}594 non-compiling mutants; hashing compiled images---the GPU analogue of
Trivial Compiler Equivalence \citep{papadakis2015tce}---removes 456
duplicates and host-only no-ops; tiny-input probes, a watchdog, and fork-level
isolation quarantine crashers and hangs. Table~\ref{tab:stats} gives the
pipeline in numbers. \emph{Scoring:} a protocol's score is
the fraction of the 7{,}384 witnessed mutants it kills under the benchmark's
own oracle ($\texttt{allclose}$, $\texttt{atol}{=}\texttt{rtol}{=}10^{-2}$
unless the protocol specifies otherwise). 
% Witnesses ship with the artifact, and every number below is reproducible from its journals.

\begin{table}[t]
\centering
\caption{The mutation pipeline of Fig.~\ref{fig:pipeline} in numbers, at
operator scale (levels 1--2) and architecture scale (level 3).
``Distinct'' counts mutants surviving compilation, compiled-image
deduplication, and equivalence filtering; ``witnessed'' additionally
requires a validity-gated input on which the mutant verifiably fails. The
124 rules split boundary 39, precision 30, semantic 22, synchronization 14,
indexing 14, arithmetic 5; 96 fire with a witness (Table~\ref{tab:tax}).
Substrates total 44{,}270 lines of CUDA across 360 device kernels; at
operator scale a problem carries 37 mutants at the median (199 at most) and
5 survivors (81 at most). The kill matrix comprises over 120{,}000
mutant--input evaluations, each mutant compiled in 84\,ms by NVRTC versus
200\,s through the torch extension path.}
\label{tab:stats}
\vspace{5pt}
\small
\begin{tabular}{@{}lrr@{}}
\toprule
& Operators (levels 1--2) & Architectures (level 3) \\
\midrule
Problems (substrates)               & 189 & 48 \\
Mutation rules                      & 124 & 124 \\
\midrule
Mutants generated                   & 10{,}303 & 12{,}589 \\
\quad $-$ non-compiling           & 1{,}594 & 38 \\
\quad $-$ duplicate compiled image & 279 & 155 \\
\quad $-$ equivalent to original  & 177 & 35 \\
Distinct mutants                    & 8{,}253 & 12{,}361 \\
\quad \textbf{of which witnessed}   & \textbf{7{,}384} & \textbf{8{,}519} \\
\midrule
Witnessed mutants killed by the official inputs & 6{,}136 & 7{,}043 \\
\textbf{Witnessed mutants missed by the official inputs} & \textbf{1{,}248} & \textbf{1{,}476} \\
Miss rate                           & 16.9\% & 17.3\% \\
\bottomrule
\end{tabular}
\end{table}

\section{How strong is the official check?}
\label{sec:measurement}

\begin{figure}[t]
\centering
\includegraphics[width=\linewidth]{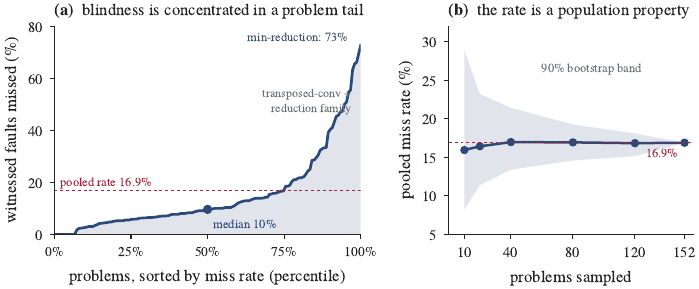}
\caption{Population structure of the official check's blindness at operator
scale. \textbf{(a)}~Per-problem miss rate, sorted: half the problems lose
under 10\% of witnessed faults, while a tail of transposed-convolution and
reduction problems---exactly the shapes tolerance vacuity predicts---lose
40--73\%. \textbf{(b)}~Pooled miss rate under bootstrap subsampling of
problems: the 90\% band contracts onto 16.9\%, so the headline is a
property of the benchmark, not of our sample.}
\label{fig:population}
\end{figure}

\paragraph{Headline.} The official protocol---each problem's own
\texttt{get\_inputs()}, five seeds---detects 6{,}136 of the 7{,}384 witnessed
faults: 83.1\%. \textbf{One in six provably-detectable faults survives}
(1{,}248). The pool behind these rates is summarized in
Table~\ref{tab:stats}: 124 rules fire 10{,}303 times across 44{,}270 lines
of substrate CUDA, and over 120{,}000 mutant--input evaluations build the
kill matrix. Bootstrap resampling over problems shows the pooled rate is a
population property, not sampling noise: the 90\% band contracts from
$[8,29]$\% at ten problems onto 16.9\% at the full set
(Fig.~\ref{fig:population}b). The rate's structure matters as much as its
value: the per-problem distribution is heavily right-tailed
(Fig.~\ref{fig:population}a)---the median problem loses 10\% of its
witnessed faults while a tail dominated by transposed-convolution and
reduction problems loses 40--73\%---and where the misses live by fault
family is the paper's central empirical fact (Fig.~\ref{fig:mechanisms}a,
Table~\ref{tab:tax}; per-operator detail in Appendix~\ref{app:ops}).

\begin{table}[t]
\centering
\caption{Operator taxonomy and where the official check fails. Witnessed
mutants per family and the share the official five-input protocol misses.
The gradient is the paper's central empirical fact: textbook arithmetic
mutations are mostly caught, while the implementation-level families real
kernel bugs occupy---boundary, synchronization, precision---escape at
$3\times$ to $9\times$ the rate.}
\label{tab:tax}
\vspace{5pt}
\setlength{\tabcolsep}{4.5pt}
\begin{tabular}{lrrrl}
\toprule
\textbf{Family} & \textbf{Rules} & \textbf{Witnessed} & \textbf{Missed} & \textbf{Example operators} \\
\midrule
Arithmetic & 5 & 3{,}349 & 8.7\% & \emph{$+\!\to\!-$, $*\!\to\!/$, off-by-one constant} \\
Indexing & 13 & 749 & 14.4\% & \emph{axis swap, stride confusion, transposed access} \\
Semantic & 18 & 858 & 14.9\% & \emph{tie-breaking, fused-op reordering, flag flips} \\
Boundary & 27 & 1{,}735 & 22.9\% & \emph{guard deletion, ceil$\to$floor grid, tail drop} \\
Synchronization & 6 & 400 & 27.8\% & \emph{barrier removal, \texttt{\_\_syncwarp} weakening} \\
Precision & 27 & 271 & \textbf{78.6\%} & \emph{fp16 accumulators, no-stabilizer, fast-math} \\
\midrule
All & 96 & 7{,}362 & 16.9\% & \\
\bottomrule
\end{tabular}
\end{table}

\paragraph{Where the misses live.} Table~\ref{tab:tax} breaks the 16.9\% down
by fault family, and the gradient is stark. Textbook mutations---the operator
swaps that dominate classical mutation testing---are caught at 91\%: random
dense inputs excite arithmetic everywhere, so arithmetic faults have nowhere
to hide. The families that escape are precisely the ones real GPU bugs
inhabit: \emph{boundary} faults (22.9\% missed) hide because official shapes
are aligned and remainder blocks never execute; \emph{synchronization} faults
(27.8\%) hide because small aligned workloads rarely lose the race;
\emph{precision} faults (78.6\%) hide because the tolerance forgives them by
construction. A checker validated on hand-seeded arithmetic bugs would look
excellent and be blind where it matters.

\paragraph{Mechanisms, quantified.} Three regularities organize the
survivors. \emph{Vacuity scales with output magnitude:} softmax contributes 28
tolerance-blind survivors where structurally identical log-softmax contributes
9---log-domain outputs are $O(10)$, collapsing the blind band of
Fig.~\ref{fig:mechanisms}b. \emph{The ceiling caps input aggressiveness:} the
obvious fix, ``test with harsher inputs,'' is unsound past the measured
boundary of Fig.~\ref{fig:mechanisms}b; our targeted suites therefore use
same-sign magnitudes. \emph{Complementarity:} 90 mutants are killed
\emph{only} by dense random inputs---under sparse or spiky inputs a wrong
index usually reads another zero---so targeted inputs complement rather than
replace the official distribution. The empirically optimal suite shape is one
dense-random input plus one or two structure-targeted ones, which is exactly
what the optimizer of \S\ref{sec:synthesis} discovers.

\paragraph{Adjudication.} Of the mutants surviving every input known at
screening time, LLM-generated targeted inputs later produced witnesses for
431 (\S\ref{sec:synthesis}); the remaining never-witnessed mutants are
equivalence or oracle-blindness candidates and stay outside the witnessed
denominator---charged to no protocol.

\section{Auditing hardened checkers}
\label{sec:audit}

A metric that can score the official protocol can score its proposed
replacements. We re-implement KernelBench-Verified from its released source:
four deterministic scalings of each problem's own inputs
($\times 1$, $\times 3$, $\times 0.01$, $\times{-1}$; shapes never varied;
integer tensors never scaled) plus an fp32 tolerance of $10^{-3}$. We
reconstruct the Correctness-Illusion-style fuzzer from its paper (log-uniform
magnitudes spanning six decades; no code is released). All protocols are
scored on a unified denominator---8{,}215 witnessed mutants across 235
operator- and architecture-level problems---and every protocol includes the
original distribution.

\begin{figure}[t]
\centering
\includegraphics[width=0.66\linewidth]{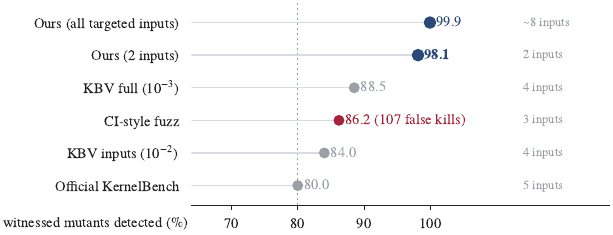}
\caption{Protocol audit on the unified denominator (8{,}215 witnessed
mutants across 235 operator- and architecture-level problems). ``Ours'' is
the official dense-random input plus one per-problem input chosen by set
cover (\S\ref{sec:synthesis}). Hardened protocols improve on the official
check (dashed line) but plateau below optimized selection at half the input
budget; the fuzz baseline's score comes with 107 false rejections of correct
kernels. Fig.~\ref{fig:matrix} dissects the blind band into its
miss-patterns.}
\label{fig:audit}
\end{figure}

\begin{figure}[t]
\centering
\includegraphics[width=\linewidth]{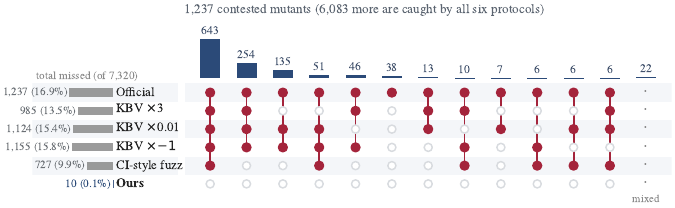}
\caption{Anatomy of the blind band: the kill matrix collapsed to its
distinct \emph{miss-patterns} (UpSet form). Each column is one pattern of
``which protocols miss this mutant'' (filled = missed), its bar the number
of operator-scale mutants with that pattern; left margins give each
protocol's total misses (cf.\ Fig.~\ref{fig:audit}). The band's core is
the leftmost column: 643 mutants that every baseline misses and only our
targeted suite detects. Each KBV scaling removes a different, partially
overlapping slice (columns 2--5); the fuzzer removes more; our suite misses
10 of 7{,}320 in the residual mixed tail.}
\label{fig:matrix}
\end{figure}

\paragraph{Decomposing KBV.} Of KernelBench-Verified's $+8.5$ points over the
official protocol, $+4.0$ come from the four hidden distributions and $+4.5$
from the tighter tolerance: \textbf{the tolerance change outweighs all four
designed distributions combined}. Their paper reports the same asymmetry from
the outside---hidden tests explain a 15\% speedup drop at level 1 but only 3\%
at level 2---and states that it ``does not quantify how many failures were
solely attributable to tighter tolerance versus distributional mismatches.''
The kill matrix computes precisely this.

\paragraph{Why constant scalings plateau.} All four KBV transforms rescale
magnitude; none varies shape or structure. Remainder-block boundary faults and
index-arithmetic faults are therefore unreachable \emph{by construction},
independent of how many scaled configurations are added---exactly the
families Table~\ref{tab:tax} shows the official check already misses most.
This is the kind of blind spot invisible to anecdote-based validation and
obvious under a metric.

\paragraph{The fuzz baseline crosses the ceiling.} The reconstructed fuzzer
buys its 86.2\% partly with invalid inputs: at the top of its magnitude range
it rejected \emph{correct} kernels 107 times in our audit---the
Fig.~\ref{fig:mechanisms}b failure, committed by a published recipe. Detection
bought with invalid inputs is not detection; a deployed benchmark would be
rejecting honest submissions.

\paragraph{A cautionary replication note.} An early version of our own audit
guessed $\mathcal{N}(0,1)\times10^{4}$ for KBV's large-magnitude distribution
and observed false rejections of correct kernels; the fault was our guess, not
their protocol---their published $\times 3$ cannot cross the ceiling. We keep
the episode on record because it argues the thesis better than any experiment
we designed: without a measured validity gate, an input designer---human or
model---cannot tell when they have crossed it.

\section{From measurement to synthesis}
\label{sec:synthesis}

\begin{figure}[t]
\centering
\includegraphics[width=\linewidth]{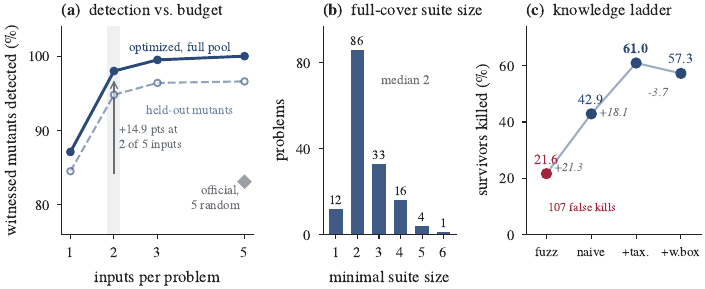}
\caption{Synthesis results. \textbf{(a)}~Detection versus input budget:
greedy set cover on the kill matrix, scored on the full pool (solid) and on
held-out mutants under dev-half selection (dashed). \textbf{(b)}~Minimal
full-coverage suite size per problem: median two inputs, maximum six.
\textbf{(c)}~Killing power of generated suites on 1{,}154 official-input
survivors as generator knowledge grows (Table~\ref{tab:ladder}): the
measurement-derived taxonomy beats both ignorance and white-box fault
listings.}
\label{fig:synthesis}
\end{figure}

\paragraph{Suites as set cover.} Once the kill matrix exists, suite
construction is the classical covering problem
\citep{harrold1993methodology,yoo2012regression}. Greedy cover reaches full
coverage of witnessed mutants with a median of \textbf{two} inputs per
problem (Fig.~\ref{fig:synthesis}b): one dense-random plus one targeted input
suffice for 86 of the 152 problems with a non-trivial witnessed pool, and no
problem needs more than six. The
budget curve is steep (Table~\ref{tab:budget}, Fig.~\ref{fig:synthesis}a):
at the official five-input budget an optimized suite detects every
witnessed mutant, against the official 83.1\%. The gain is \emph{chosen} testing,
not more.

\begin{table}[t]
\centering
\caption{Detection versus input budget on the witnessed level-1/2 pool.
Held-out: suites selected on a 50\% dev split of each problem's mutants and
scored only on the other half.}
\label{tab:budget}
\vspace{5pt}
\begin{tabular}{lccccc}
\toprule
& \multicolumn{4}{c}{Optimized, budget $b$} & Official \\
\cmidrule(lr){2-5}
Scored on & $b{=}1$ & $b{=}2$ & $b{=}3$ & $b{=}5$ & 5 random \\
\midrule
Full pool & 87.1\% & \textbf{98.0\%} & 99.5\% & 100.0\% & 83.1\% \\
Held-out mutants & 84.5\% & 94.8\% & 96.4\% & 96.6\% & 82.9\% \\
\bottomrule
\end{tabular}
\end{table}

\paragraph{Holdout validation.} To rule out overfitting the selection to the
measured pool, we split each problem's witnessed mutants 50/50 by id-hash,
select on the dev half only, and score on the test half
(Table~\ref{tab:budget}, bottom row). Dev-selected suites detect 94.8\% of
held-out mutants at budget two---an ${\sim}3$-point generalization gap---
indicating the chosen inputs capture fault \emph{families}, not memorized
individuals, consistent with the mechanism analysis of
\S\ref{sec:measurement}.

\paragraph{What must a generator know?} Our targeted suites were written by an
LLM given three artifacts of the measurement: the blind-spot taxonomy, a
per-family attack playbook, and the ceiling of Fig.~\ref{fig:mechanisms}b.
They killed 431 previously-unkilled mutants at a 94\% suite-validity rate---
and the 6\% of suites the gate rejected were precisely ceiling violations. To
isolate how much of this is the metric's contribution, we fix a hard
target---1{,}154 mutants that survive all official inputs, across 30
problems---and vary only the generator's knowledge (Table~\ref{tab:ladder},
Fig.~\ref{fig:synthesis}c).

\begin{table}[t]
\centering
\caption{The knowledge ladder: identical task (kill 1{,}154 official-input
survivors), increasing generator knowledge. Validity = share of generated
suites that do not reject the correct kernel. The taxonomy rung accumulated
more suites per problem than the single-shot rungs, so its edge over the
naive prompt is an upper bound; the ordering is unaffected.}
\label{tab:ladder}
\vspace{5pt}
\begin{tabular}{llcc}
\toprule
Rung & Generator knows & Survivors killed & Validity \\
\midrule
R0\; random fuzz & nothing & 21.6\% & 107 false kills \\
R1\; naive prompt & ``tests may miss subtle bugs'' & 42.9\% & 60/60 \\
R2\; $+$ taxonomy \& ceiling & fault families, safe magnitudes & \textbf{61.0\%} & 94\% (accum.) \\
R3\; $+$ white-box diffs & the mutated source sites & 57.3\% & 55/56 \\
\bottomrule
\end{tabular}
\end{table}

Two readings matter. First, each increment of \emph{measurement-derived}
knowledge buys detection; the taxonomy nearly triples the fuzz baseline
($21.6\%\to61.0\%$). Second, white-box access to the faults themselves does
\emph{not} beat the taxonomy: shown ten concrete mutants, the generator
overfits its inputs to them, while family-level knowledge generalizes to the
whole pool. The measurement's abstraction is worth more than its raw
instances---which is what makes these prompted rungs a credible floor for a
learned generator whose reward is the kill rate computed here.

\section{Does depth hide faults?}
\label{sec:depth}

Level-3 KernelBench problems are whole architectures. Their substrates are
multi-kernel pipelines---one \texttt{\_\_global\_\_} per layer, 2 to 140
kernels---so every mutant carries a layer tag, and a question inaccessible at
operator scale becomes measurable: \emph{does an early fault reach the
network output?}

\begin{table}[t]
\centering
\caption{Operator scale versus architecture scale. Distinct = behaviourally
distinct compilable mutants; survival = share not killed by the official
inputs; witnessed = mutants with a validity-gated kill witness. The level-3
witness search is far shallower (2--4 targeted suites per problem versus the
full matrix at operator scale), so its witnessed miss rate is a conservative
lower bound and its distinct-survival an upper bound.}
\label{tab:levels}
\vspace{5pt}
\begin{tabular}{lcccc}
\toprule
Scale & Problems & Distinct & Survival (distinct) & Missed (witnessed) \\
\midrule
Operators (levels 1--2) & 189 & 8{,}253 & 25.7\% & 16.9\% \\
Architectures (level 3) & 48 & 12{,}361 & \textbf{43.0\%} & $\ge$\,17.3\% \\
\bottomrule
\end{tabular}
\end{table}

\begin{figure}[t]
\centering
\includegraphics[width=\linewidth]{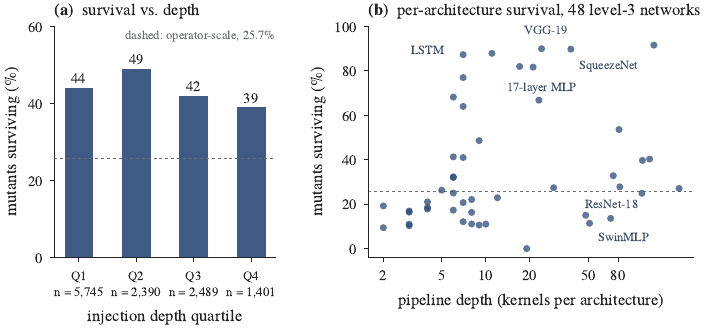}
\caption{Architecture-scale results (distinct-mutant accounting).
\textbf{(a)}~Survival by injection-depth quartile across 48 architectures:
faults with more network downstream survive more. \textbf{(b)}~Per-architecture
survival versus pipeline depth: deep homogeneous pipelines (VGG, LSTM, MLP
chains) approach opacity, while normalization- and branch-dense networks
(ResNet-18, SwinMLP) remain checkable at any depth.}
\label{fig:depth}
\end{figure}

\textbf{Architecture-level checking is weaker on every accounting}
(Table~\ref{tab:levels}). Under strict witnessed accounting, 17.3\% of
architecture-level faults escape the official inputs against 16.9\% at
operator scale---and this is a floor, since the level-3 witness search is far
shallower. The distinct-mutant upper bound is $1.7\times$: 43.0\% versus
25.7\%. More telling than either aggregate is the structure. Per architecture
(Fig.~\ref{fig:depth}b), survival splits sharply: deep \emph{homogeneous}
pipelines are near-opaque to the official check---VGG-19 90\%, SqueezeNet
90\%, LSTM stacks 77--87\%, a 17-layer MLP 82\%---while architectures dense
in normalization and branching stay comparatively transparent even at extreme
depth (ResNet-18 at 51 kernels: 11\%; SwinMLP at 71 kernels: 14\%); the
per-architecture median of 25.7\% matches the operator scale exactly, so the
aggregate gap is carried entirely by the homogeneous tail. Depth supplies the
opportunity for masking; homogeneity---long chains without
renormalization---realizes it. Within networks the same physics appears
(Fig.~\ref{fig:depth}a): faults injected in the front half survive at
44--49\%, falling to 39\% in the output quartile, a modest but consistent
gradient ($n=1{,}401$--$5{,}745$ per quartile). The direction of travel for
the field---from single operators toward end-to-end generated models---is
precisely the direction in which its oracles weaken.

\paragraph{Two problems no oracle can referee.} The two level-3 problems our
admission gate could never pass turn out to be unpassable in principle. For
\texttt{48\_Mamba2ReturnY}, whose reference exponentiates cumulative sums of
unbounded random parameters, outputs reach $10^{20}$ and the official fp32
forward violates the benchmark tolerance against its own fp64 evaluation at
352 positions; for \texttt{45\_UNetSoftmax}, whose blocks chain softmax into
batch normalization---a variance amplifier---the official fp32 forward
deviates from fp64 by up to $0.74$ (9{,}379 violations), \emph{farther than
our rejected candidate sits from the reference}. No fp32 implementation,
including the reference itself, can be adjudicated on these problems. No
prior audit noticed: KernelBench-Verified ships hidden tests for both (its
validity filter catches only NaN/Inf, and finite $10^{20}$ outputs pass);
robust-kbench's filters never touch level 3; the instability of the naive
Mamba segment-sum is acknowledged upstream (the reference implementation
names it \texttt{segsum\_unstable}) but had not been connected to the
benchmark. Scores reported on these two problems are noise---a concrete
instance of what patching without measuring cannot see.

\section{Related work}
\label{sec:related}

\paragraph{Kernel benchmarks and their checkers.} KernelBench
\citep{ouyang2025kernelbench} established the task and the randomized-%
\texttt{allclose} check we audit; TritonBench \citep{li2025tritonbench} and
successors broaden the task surface. That the check is gameable is documented from
both sides: Kevin \citep{baronio2025kevin} reports reward hacking during RL
against it, while KernelBench-Verified \citep{zhang2026kbverified} finds
hard-coded bypasses and hardens the protocol, as do robust-kbench
\citep{lange2025robust} and The Correctness Illusion
\citep{sarkar2026illusion}. All of these patch or exploit the oracle; none
measures it. Our metric is the missing instrument, and \S\ref{sec:audit} is
what it reads on their patches.

\paragraph{Auditing benchmark oracles elsewhere.} EvalPlus
\citep{liu2023evalplus} showed HumanEval's tests pass wrong programs and that
augmenting them reorders model rankings; EvalPerf \citep{liu2024evalperf}
extends this to efficiency. Concurrently with us, program-variant test augmentation \citep{li2026sting}
stresses SWE-bench Verified with semantically altered patches and GateTruth
\citep{bhadra2026gatetruth} mutation-audits RTL benchmark testbenches. None of
these domains has a graded numerical oracle, so none confronts tolerance
vacuity or a validity ceiling---the two mechanisms that make the kernel
setting distinct (\S\ref{sec:method}) and that our audit shows existing
kernel-domain patches mishandle.

\paragraph{Mutation testing, including on GPU code.} Mutation analysis begins
with \citet{demillo1978hints}; \citet{jia2011survey} survey it;
\citet{offutt1996sufficient} ground operator selection;
\citet{papadakis2015tce} give the equivalent-mutant filter whose CUBIN-hash
analogue we use. On GPU code, MUTGPU \citep{zhu2020mutgpu} defined
CUDA-specific operators and CLTestCheck \citep{peng2019cltestcheck} seeded
faults to score OpenCL test suites; \citet{bradbury2006concurrent} pioneered
concurrency operators; GPUVerify \citep{betts2012gpuverify} and test
amplification \citep{leung2012amplification} offer verification-flavoured
alternatives. These score developer suites on hand-written kernels; none
targets a benchmark oracle, and none gates input validity against
floating-point variance.

\paragraph{Synthesis, compilers, numerics.} Suite minimization as set cover
goes back to \citet{harrold1993methodology,yoo2012regression};
mutation-guided test generation is deployed industrially
\citep{foster2025meta}. Compiler-testing lines generate \emph{equivalent}
variants to expose miscompilation---EMI \citep{le2014emi}, NNSmith
\citep{liu2023nnsmith}---where we generate \emph{inequivalent} variants to
expose weak checkers. The floating-point non-associativity underlying our
ceiling is characterized by \citet{shanmugavelu2024fpna} and
\citet{goldberg1991floating}; we measure where it bites and promote it to a
hard constraint on test generation. Nearest to a mutation--performance
crossover, MUPPET \citep{miao2024muppet} mutates OpenMP directives to
\emph{search for} speedups---the same machinery aimed at the opposite
problem. The combination needed here---mutation analysis grading the oracle
of a kernel benchmark---did not previously exist.

\section{Limitations}
\label{sec:limitations}

\paragraph{Adequacy is relative to a fault model.} Our 124 rules over naive
substrates cannot express faults living in structures we do not generate---
tensor-core paths, double-buffered pipelines, some warp-level idioms---nor
multi-site interactions. The metric licenses \emph{comparative} claims (suite
A misses faults suite B catches) and \emph{existence} claims (this protocol
misses these witnessed faults); it cannot certify a passing kernel correct,
and we never use it to.

\paragraph{Realism probe.} As a first direct check on the fault model, we
prompted an LLM to write leaderboard-style \emph{optimized} kernels (tiling,
\texttt{float4}, warp shuffles; no taxonomy shown) for 60 random problems and
gated them against the official reference. Three were incorrect: a compile
error; a
misaligned-address crash from an unguarded vectorized load---squarely in our
boundary/guard family; and a value error in a fused GEMM--GroupNorm, matching
our accumulation/semantic families. The bug sample is small, but both runtime
bugs fall inside the taxonomy. The exercise also exposed a harness blind spot
that only realistic style triggers
(\texttt{\_\_launch\_bounds\_\_}-qualified kernels defeating naive
\texttt{extern "C"} injection)---the checker-measurement lesson applied to our
own tooling.

\paragraph{Scope.} Substrates are LLM-authored and gate-verified rather than
sampled from submissions; results are from one GPU generation (H100). The
level-3 study covers 48 of 50 architectures---the remaining two are excluded
on proof that no fp32 implementation can be adjudicated on them
(\S\ref{sec:depth})---and its witness search is shallower than at operator
scale, which Table~\ref{tab:levels} treats as a one-sided bound. The
tolerance is fixed at the benchmark's own $10^{-2}$ by design, with
\S\ref{sec:audit} isolating the effect of changing it.

\section{Conclusion}
\label{sec:conclusion}

Benchmarks for generated GPU kernels have been patching their correctness
checkers blind. Mutation analysis, adapted to graded numerical oracles---a
witnessed denominator, a measured validity ceiling, tractable compilation---
turns checker quality into a number, explains the number through two
mechanisms and a six-family taxonomy, prices existing patches, and converts
suite design into optimization that generalizes to held-out faults. The
measurement also yields its most useful by-product for what comes next: a
fault taxonomy that teaches a test generator more than the faults themselves.
We release the pool, witnesses, suites, and pipeline as
\textbf{KernelBench-M}, so that the next patch to a kernel benchmark can ship
with its coverage measured rather than asserted.

\section*{Reproducibility statement}
Every number in the paper is computed from journals shipped with the
KernelBench-M artifact: the 124 mutation rules (Appendix~\ref{app:rules}),
the gate-verified CUDA substrates, per-mutant kill records with the suite
that witnesses each kill, the dev/test split of \S\ref{sec:synthesis}, and
the pipeline scripts that regenerate the kill matrix, the protocol audit,
and the set-cover suites from those records. Figures are produced by the
scripts in the repository from the released JSON summaries. All experiments
ran on a single NVIDIA H100 (80\,GB, driver 535.161.08) with PyTorch~2.7.1
(CUDA~12.6 build) and NVRTC~12.6; the measurement campaign used
approximately 30 GPU-hours.

\section*{Ethics statement}
This work audits the correctness checkers of public benchmarks; it involves
no human subjects, personal data, or deployed systems. Its findings make
benchmark verdicts harder to game and are intended to reduce, not enable,
reward hacking in kernel-generating models. The audited benchmarks and the
patches we re-implement are cited and used under their published terms; we
report their weaknesses with the reconstruction details needed to check our
claims and note where an earlier version of our own audit was at fault
(\S\ref{sec:audit}).

\section*{The use of large language models}
LLMs are part of the measured system, not only a writing aid: substrates
are LLM-authored CUDA implementations admitted by an automated gate
(\S\ref{sec:method}), the fine-grained mutation families were mined with LLM
assistance from the rule brief in the artifact, the targeted input suites
and the knowledge-ladder rungs of \S\ref{sec:synthesis} are LLM-generated
under the prompts reproduced in Appendix~\ref{app:ladder}, and the realism
probe of \S\ref{sec:limitations} uses LLM-written optimized kernels.
Every LLM-authored artifact was produced with the OpenAI Codex CLI
(model \texttt{gpt-5.6-sol}) through one non-interactive \texttt{codex exec}
call per task with a 900\,s timeout, sampled once and retried only on
failure; reasoning effort was left at its default (\texttt{none}) except for
eight substrate-repair runs at \texttt{high}. No LLM output entered the paper unverified: every
substrate passed the admission gate, every generated suite passed the
validity gate before it could kill a mutant, and all numbers derive from the
released journals. LLMs were also used to polish the prose.

\bibliography{references}

@article{demillo1978hints,
  author={DeMillo, Richard A. and Lipton, Richard J. and Sayward, Frederick G.},
  title={Hints on Test Data Selection: Help for the Practicing Programmer},
  journal={IEEE Computer}, volume={11}, number={4}, pages={34--41}, year={1978}}

@article{jia2011survey,
  author={Jia, Yue and Harman, Mark},
  title={An Analysis and Survey of the Development of Mutation Testing},
  journal={IEEE Transactions on Software Engineering}, volume={37}, number={5},
  pages={649--678}, year={2011}}

@article{offutt1996sufficient,
  author={Offutt, A. Jefferson and Lee, Ammei and Rothermel, Gregg and Untch, Roland H. and Zapf, Christian},
  title={An Experimental Determination of Sufficient Mutant Operators},
  journal={ACM Transactions on Software Engineering and Methodology}, volume={5}, number={2}, pages={99--118}, year={1996}}

@inproceedings{papadakis2015tce,
  author={Papadakis, Mike and Jia, Yue and Harman, Mark and Le Traon, Yves},
  title={Trivial Compiler Equivalence: A Large Scale Empirical Study of a Simple, Fast and Effective Equivalent Mutant Detection Technique},
  booktitle={Proceedings of the 37th International Conference on Software Engineering (ICSE)}, pages={936--946}, year={2015}}

@article{harrold1993methodology,
  author={Harrold, Mary Jean and Gupta, Rajiv and Soffa, Mary Lou},
  title={A Methodology for Controlling the Size of a Test Suite},
  journal={ACM Transactions on Software Engineering and Methodology}, volume={2}, number={3}, pages={270--285}, year={1993}}

@article{yoo2012regression,
  author={Yoo, Shin and Harman, Mark},
  title={Regression Testing Minimization, Selection and Prioritization: A Survey},
  journal={Software Testing, Verification and Reliability}, volume={22}, number={2}, pages={67--120}, year={2012}}

@inproceedings{zhu2020mutgpu,
  author={Zhu, Qianqian and Zaidman, Andy},
  title={Massively Parallel, Highly Efficient, but What About the Test Suite Quality? Applying Mutation Testing to {GPU} Programs},
  booktitle={IEEE International Conference on Software Testing, Validation and Verification (ICST)}, pages={209--219}, year={2020}}

@inproceedings{peng2019cltestcheck,
  author={Peng, Chao and Rajan, Ajitha},
  title={{CLTestCheck}: Measuring Test Effectiveness for {GPU} Kernels},
  booktitle={Fundamental Approaches to Software Engineering (FASE)}, series={LNCS}, volume={11424}, pages={315--331}, year={2019}}

@inproceedings{bradbury2006concurrent,
  author={Bradbury, Jeremy S. and Cordy, James R. and Dingel, Juergen},
  title={Mutation Operators for Concurrent {Java} ({J2SE} 5.0)},
  booktitle={Second Workshop on Mutation Analysis (Mutation)}, year={2006}}

@inproceedings{betts2012gpuverify,
  author={Betts, Adam and Chong, Nathan and Donaldson, Alastair F. and Qadeer, Shaz and Thomson, Paul},
  title={{GPUVerify}: A Verifier for {GPU} Kernels},
  booktitle={ACM SIGPLAN Conference on Object-Oriented Programming, Systems, Languages, and Applications (OOPSLA)}, pages={113--132}, year={2012}}

@inproceedings{leung2012amplification,
  author={Leung, Alan and Gupta, Manish and Agarwal, Yuvraj and Gupta, Rajesh and Jhala, Ranjit and Lerner, Sorin},
  title={Verifying {GPU} Kernels by Test Amplification},
  booktitle={ACM SIGPLAN Conference on Programming Language Design and Implementation (PLDI)}, pages={383--394}, year={2012}}

@inproceedings{liu2023evalplus,
  author={Liu, Jiawei and Xia, Chunqiu Steven and Wang, Yuyao and Zhang, Lingming},
  title={Is Your Code Generated by {ChatGPT} Really Correct? Rigorous Evaluation of Large Language Models for Code Generation},
  booktitle={Advances in Neural Information Processing Systems (NeurIPS)}, volume={36}, year={2023}}

@inproceedings{liu2024evalperf,
  author={Liu, Jiawei and Xie, Songrun and Wang, Junhao and Wei, Yuxiang and Ding, Yifeng and Zhang, Lingming},
  title={Evaluating Language Models for Efficient Code Generation},
  booktitle={Conference on Language Modeling (COLM)}, year={2024}}

@inproceedings{ouyang2025kernelbench,
  author={Ouyang, Anne and Guo, Simon and Arora, Simran and Zhang, Alex L. and Hu, William and R{\'e}, Christopher and Mirhoseini, Azalia},
  title={{KernelBench}: Can {LLMs} Write Efficient {GPU} Kernels?},
  booktitle={Proceedings of the 42nd International Conference on Machine Learning (ICML)},
  series={PMLR}, volume={267}, pages={47356--47415}, year={2025}}

@article{zhang2026kbverified,
  author={Zhang, Yunxiang and Yu, Ping and Wang, Jianyu and Fan, Max (Xiangjun) and Reed, Julian and Mirhoseini, Azalia and Su, Will},
  title={{KernelBench-Verified}: Do {LLM}-Generated Kernels Actually Beat {PyTorch}?},
  journal={arXiv preprint arXiv:2607.16241}, year={2026}}

@article{sarkar2026illusion,
  author={Sarkar, Dipankar},
  title={The Correctness Illusion in {LLM}-Generated {GPU} Kernels},
  journal={arXiv preprint arXiv:2606.20128}, year={2026}}

@article{lange2025robust,
  author={Lange, Robert Tjarko and Sun, Qi and Prasad, Aaditya and Faldor, Maxence and Tang, Yujin and Ha, David},
  title={Towards Robust Agentic {CUDA} Kernel Benchmarking, Verification, and Optimization},
  journal={arXiv preprint arXiv:2509.14279}, year={2025}}

@inproceedings{li2025tritonbench,
  author={Li, Jianling and Li, Shangzhan and Gao, Zhenye and Shi, Qi and Li, Yuxuan and Wang, Zefan and Huang, Jiacheng and Wang, Haojie and Wang, Jianrong and Han, Xu and Liu, Zhiyuan and Sun, Maosong},
  title={{TritonBench}: Benchmarking Large Language Model Capabilities for Generating {Triton} Operators},
  booktitle={Findings of the Association for Computational Linguistics: ACL 2025}, pages={23053--23066}, year={2025}}

@article{baronio2025kevin,
  author={Baronio, Carlo and Marsella, Pietro and Pan, Ben and Guo, Simon and Alberti, Silas},
  title={Kevin: Multi-Turn {RL} for Generating {CUDA} Kernels},
  journal={arXiv preprint arXiv:2507.11948}, year={2025}}

@article{li2026sting,
  author={Li, Chenglin and Xu, Yisen and Wang, Zehao and Tan, Shin Hwei and Chen, Tse-Hsun},
  title={Probe to Generate: Program Variant-Guided Test Augmentation for Repository-Level Repair Benchmarks},
  journal={arXiv preprint arXiv:2604.01518 (to appear, ASE)}, year={2026}}

@article{bhadra2026gatetruth,
  author={Bhadra, Meet},
  title={{GateTruth}: Auditing the Rigor of {RTL} Design Benchmarks via Mutation Testing},
  journal={arXiv preprint arXiv:2608.12635}, year={2026}}

@inproceedings{foster2025meta,
  author={Harman, Mark and Ritchey, Jillian and Harper, Inna and Sengupta, Shubho and Mao, Ke and Gulati, Abhishek and Foster, Christopher and Robert, Herv{\'e}},
  title={Mutation-Guided {LLM}-based Test Generation at {Meta}},
  booktitle={Companion Proceedings of the 33rd ACM International Conference on the Foundations of Software Engineering (FSE Companion)},
  pages={180--191}, year={2025}}

@inproceedings{liu2023nnsmith,
  author={Liu, Jiawei and Lin, Jinkun and Ruffy, Fabian and Tan, Cheng and Li, Jinyang and Panda, Aurojit and Zhang, Lingming},
  title={{NNSmith}: Generating Diverse and Valid Test Cases for Deep Learning Compilers},
  booktitle={International Conference on Architectural Support for Programming Languages and Operating Systems (ASPLOS)}, pages={530--543}, year={2023}}

@inproceedings{le2014emi,
  author={Le, Vu and Afshari, Mehrdad and Su, Zhendong},
  title={Compiler Validation via Equivalence Modulo Inputs},
  booktitle={ACM SIGPLAN Conference on Programming Language Design and Implementation (PLDI)}, pages={216--226}, year={2014}}

@inproceedings{miao2024muppet,
  author={Miao, Dolores and Laguna, Ignacio and Georgakoudis, Giorgis and Parasyris, Konstantinos and Rubio-Gonz{\'a}lez, Cindy},
  title={{MUPPET}: Optimizing Performance in {OpenMP} via Mutation Testing},
  booktitle={Workshop on Programming Models and Applications for Multicores and Manycores (PMAM@PPoPP)}, pages={22--31}, year={2024}}

@inproceedings{shanmugavelu2024fpna,
  author={Shanmugavelu, Sanjif and Taillefumier, Mathieu and Culver, Christopher and Hernandez, Oscar R. and Coletti, Mark and Sedova, Ada},
  title={Impacts of Floating-Point Non-Associativity on Reproducibility for {HPC} and Deep Learning Applications},
  booktitle={SC24-W: Workshops of the International Conference for High Performance Computing, Networking, Storage and Analysis}, pages={170--179}, year={2024}}

@article{goldberg1991floating,
  author={Goldberg, David},
  title={What Every Computer Scientist Should Know About Floating-Point Arithmetic},
  journal={ACM Computing Surveys}, volume={23}, number={1}, pages={5--48}, year={1991}}
\bibliographystyle{iclr2027_conference}

\appendix
\section{Operator inventory: where the misses come from}
\label{app:ops}

Table~\ref{tab:ops} lists the ten mutation operators contributing the most
officially-missed faults at operator scale. The two regimes visible in
Table~\ref{tab:tax} recur at operator granularity: high-volume arithmetic
operators are almost always caught (6--10\% missed), while low-volume
implementation-level operators are missed at wholesale rates---storing
through an fp16 temporary escapes the tolerance 92\% of the time, and
synchronization or boundary weakenings escape at 20--29\% because the
official aligned shapes never exercise the code they break.

\begin{table}[h]
\centering
\caption{Top mutation operators by officially-missed faults (operator scale,
witnessed accounting). ``Example site'' shows a representative mutated
source location.}
\label{tab:ops}
\vspace{5pt}
\footnotesize
\setlength{\tabcolsep}{2.5pt}
\begin{tabular}{llrrl}
\toprule
Operator & Family & Missed / witnessed & Rate & Example site \\
\midrule
\texttt{store-fp16} & precision & 169 / 183 & 92\% & \texttt{out[i] = (half)acc} \\
\texttt{mul2plus} & arithmetic & 99 / 990 & 10\% & \texttt{delta * delta} \\
\texttt{loop-start-1} & boundary & 82 / 417 & 20\% & \texttt{for (int i = 0;} \\
\texttt{mul2div} & arithmetic & 78 / 968 & 8\% & \texttt{value * norm\_weight[c]} \\
\texttt{loop-bound-minus1} & boundary & 77 / 384 & 20\% & \texttt{i < hidden\_size; i++} \\
\texttt{plus2minus} & arithmetic & 75 / 1{,}173 & 6\% & \texttt{variance + 1.0e-5} \\
\texttt{sync-remove} & sync & 58 / 198 & 29\% & \texttt{\_\_syncthreads();} \\
\texttt{sync2syncwarp} & sync & 49 / 190 & 26\% & \texttt{\_\_syncthreads();} \\
\texttt{const0to1} & semantic & 44 / 151 & 29\% & \texttt{= 0.0f;} \\
\texttt{last-row-reduction-skip} & boundary & 41 / 164 & 25\% & \texttt{for (int j = 0; j < n;} \\
\bottomrule
\end{tabular}
\end{table}

\section{Mutation-rule inventory}
\label{app:rules}
Table~\ref{tab:rules} lists every rule the released mutator applies, with the
one-line description recorded in the rule definition. Each rule is a
deterministic rewrite (a regular-expression pattern and a replacement
template) applied to the device source; every match site yields one mutant.
Rules were added in three waves: the classical and GPU-specific operators of
the base mutator, a set mined from survivor analysis, and a fine-grained set
targeting numerically subtle sites; the rule brief that guided mining ships
with the artifact.
\footnotesize
\begin{longtable}{@{}p{0.34\linewidth}p{0.62\linewidth}@{}}
\caption{The mutation-rule inventory loaded by the released mutator, by family. \textcolor{pcarm}{[TODO: the released rule files load 120 unique rules (7 definitions duplicate an existing pattern and are skipped); the main text counts 124. Reconcile.]}}\label{tab:rules}\\
\toprule \textbf{Rule} & \textbf{Description} \\ \midrule \endfirsthead
\toprule \textbf{Rule} & \textbf{Description} \\ \midrule \endhead
\bottomrule \endlastfoot
\multicolumn{2}{@{}l}{\emph{Arithmetic} (4 rules)} \\
\quad\raggedright\texttt{minus2plus} & AOR: - $\to$ + \\
\quad\raggedright\texttt{mul2div} & AOR: * $\to$ / \\
\quad\raggedright\texttt{negate-\allowbreak{}store} & UOI: negate stored expression \\
\quad\raggedright\texttt{plus2minus} & AOR: + $\to$ - \\
\addlinespace
\multicolumn{2}{@{}l}{\emph{Indexing} (14 rules)} \\
\quad\raggedright\texttt{a-\allowbreak{}row-\allowbreak{}stride-\allowbreak{}n} & use N rather than K for one matrix-A row stride \\
\quad\raggedright\texttt{argmax-\allowbreak{}final-\allowbreak{}row-\allowbreak{}alias} & Mechanism 2 (input-domain evasion): it differs only when the unique maximum is in the final row; a final-row spike kills it. \\
\quad\raggedright\texttt{argmax-\allowbreak{}final-\allowbreak{}row-\allowbreak{}value-\allowbreak{}alias} & Mechanism 2 (input-domain evasion): only the last candidate is misindexed and is rarely maximal; a final-row spike kills it. \\
\quad\raggedright\texttt{b-\allowbreak{}row-\allowbreak{}stride-\allowbreak{}k} & use K rather than N for one matrix-B row stride \\
\quad\raggedright\texttt{bdim2gdim} & blockDim $\to$ gridDim in index math \\
\quad\raggedright\texttt{bidx2bidy} & wrong block axis (single site, unlike consistent-swap) \\
\quad\raggedright\texttt{flatten-\allowbreak{}height-\allowbreak{}use-\allowbreak{}width} & substitute width for one height stride in a flattened index \\
\quad\raggedright\texttt{grid-\allowbreak{}x-\allowbreak{}use-\allowbreak{}block-\allowbreak{}y} & use the other block extent in a linear global index \\
\quad\raggedright\texttt{output-\allowbreak{}position-\allowbreak{}height-\allowbreak{}first} & decode a flattened x coordinate with the wrong extent \\
\quad\raggedright\texttt{reduction-\allowbreak{}partner-\allowbreak{}half-\allowbreak{}offset} & read the wrong partner within one tree-reduction stage \\
\quad\raggedright\texttt{row-\allowbreak{}store-\allowbreak{}stride-\allowbreak{}k} & use the reduction dimension as one output row stride \\
\quad\raggedright\texttt{smem-\allowbreak{}index-\allowbreak{}off1} & shared-memory partner index off by one \\
\quad\raggedright\texttt{tidx2tidy} & wrong thread axis \\
\quad\raggedright\texttt{warpsize-\allowbreak{}16} & reduction start half $\to$ quarter \\
\addlinespace
\multicolumn{2}{@{}l}{\emph{Semantic} (21 rules)} \\
\quad\raggedright\texttt{acc-\allowbreak{}overwrite} & accumulate $\to$ overwrite (keeps last term only) \\
\quad\raggedright\texttt{argmax-\allowbreak{}nan-\allowbreak{}policy-\allowbreak{}drop} & drop the NaN-propagation branch: differs only on NaN inputs \\
\quad\raggedright\texttt{argmax-\allowbreak{}tie-\allowbreak{}last} & argmax keeps the LAST maximum instead of the first: differs only when duplicate values exist (measure zero under rand) \\
\quad\raggedright\texttt{argmax-\allowbreak{}tie-\allowbreak{}nan-\allowbreak{}form} & argmax with NaN handling: ties now keep the LAST index; only duplicate-valued inputs differ (measure zero under rand) \\
\quad\raggedright\texttt{const0to1} & init constant 0 $\to$ 1 \\
\quad\raggedright\texttt{elu-\allowbreak{}positive-\allowbreak{}cutoff-\allowbreak{}nudge} & Mechanism 2 (input-domain evasion): positive originals take the same branch; tiny negative values kill it. \\
\quad\raggedright\texttt{fabs-\allowbreak{}simple-\allowbreak{}remove} & absolute value is an identity on the original positive corpus \\
\quad\raggedright\texttt{fmax2fmin} & max $\to$ min \\
\quad\raggedright\texttt{fmin2fmax} & min $\to$ max \\
\quad\raggedright\texttt{lower-\allowbreak{}negunit-\allowbreak{}clamp-\allowbreak{}remove} & drop a lower saturation bound \\
\quad\raggedright\texttt{nested-\allowbreak{}unit-\allowbreak{}clamp-\allowbreak{}drop-\allowbreak{}upper} & retain the lower clamp but remove the upper saturation \\
\quad\raggedright\texttt{reduce-\allowbreak{}fmax-\allowbreak{}tie-\allowbreak{}order} & fmaxf replaced by a strict comparison: identical except for NaN operands, where fmaxf ignores NaN but $>$ propagates the old value \\
\quad\raggedright\texttt{reduce-\allowbreak{}fmin-\allowbreak{}tie-\allowbreak{}order} & fminf replaced by strict comparison: NaN semantics differ \\
\quad\raggedright\texttt{relu-\allowbreak{}fmax-\allowbreak{}zero-\allowbreak{}first-\allowbreak{}remove} & remove commuted ReLU wrapper; positive inputs conceal the fault \\
\quad\raggedright\texttt{relu-\allowbreak{}fmax-\allowbreak{}zero-\allowbreak{}remove} & remove a simple ReLU wrapper; differs only for negative values \\
\quad\raggedright\texttt{relu-\allowbreak{}threshold-\allowbreak{}shift} & activation threshold nudged off zero: only tiny values differ \\
\quad\raggedright\texttt{selu-\allowbreak{}positive-\allowbreak{}cutoff-\allowbreak{}nudge} & Mechanism 2 (input-domain evasion): ordinary positive values avoid the moved cutoff; tiny negative values in [-1e-6,0] kill it. \\
\quad\raggedright\texttt{sentinel-\allowbreak{}zero} & reduction sentinel -inf $\to$ 0 (breaks all-negative rows) \\
\quad\raggedright\texttt{softsign-\allowbreak{}positive-\allowbreak{}fabs-\allowbreak{}elide} & Mechanism 2 (input-domain evasion): fabs is identical for positive originals; any negative input kills it. \\
\quad\raggedright\texttt{upper-\allowbreak{}unit-\allowbreak{}clamp-\allowbreak{}first-\allowbreak{}remove} & drop a commuted unit upper clamp \\
\quad\raggedright\texttt{upper-\allowbreak{}unit-\allowbreak{}clamp-\allowbreak{}remove} & drop a unit upper clamp, exposed by values above one \\
\addlinespace
\multicolumn{2}{@{}l}{\emph{Boundary} (38 rules)} \\
\quad\raggedright\texttt{arg-\allowbreak{}reduction-\allowbreak{}last-\allowbreak{}row-\allowbreak{}drop} & Mechanism 2 (input-domain evasion): the final row is almost never the unique random extremum; forcing the extremum into the final row kills it. \\
\quad\raggedright\texttt{ceil2floor} & ceil-div $\to$ floor-div: tail block never processed \\
\quad\raggedright\texttt{ceil2floor-\allowbreak{}lit} & (x + K-1)/K with literal K $\to$ floor \\
\quad\raggedright\texttt{cumsum-\allowbreak{}last-\allowbreak{}output-\allowbreak{}drop} & Mechanism 1 (tolerance absorption): only the last prefix output is omitted and its missing term is tiny relative to a long positive sum; a short scan or last-position spike kills it. \\
\quad\raggedright\texttt{elementwise-\allowbreak{}guard-\allowbreak{}minus1} & last element of the flat range never written: needs a spiked or misaligned tail to expose \\
\quad\raggedright\texttt{flat-\allowbreak{}n-\allowbreak{}one-\allowbreak{}extra-\allowbreak{}lane} & Mechanism 3 (shape-alignment evasion): n is block-aligned in the original elementwise case; a misaligned n kills it by exposing the one-past-end access. \\
\quad\raggedright\texttt{ge2gt} & early-return guard $\ge$ $\to$ $>$ : one OOB lane slips through \\
\quad\raggedright\texttt{grid-\allowbreak{}stride-\allowbreak{}step-\allowbreak{}double} & grid-stride loop skips every other element \\
\quad\raggedright\texttt{groupnorm-\allowbreak{}sqdev-\allowbreak{}tail-\allowbreak{}scalar-\allowbreak{}drop} & Mechanism 1 (tolerance absorption): one of many variance contributions is omitted; a high-leverage final outlier kills it. \\
\quad\raggedright\texttt{groupnorm-\allowbreak{}sum-\allowbreak{}tail-\allowbreak{}scalar-\allowbreak{}drop} & Mechanism 1 (tolerance absorption): one scalar among a huge group is absent from the mean; a dominant final scalar kills it. \\
\quad\raggedright\texttt{gt2ge} & ROR: $>$ $\to$ $\ge$ \\
\quad\raggedright\texttt{guard-\allowbreak{}drop-\allowbreak{}if} & remove single-line bounds guard entirely (silent OOB) \\
\quad\raggedright\texttt{guard-\allowbreak{}drop-\allowbreak{}ternary} & remove edge-tile ternary guard: unconditional load \\
\quad\raggedright\texttt{height-\allowbreak{}bound-\allowbreak{}use-\allowbreak{}width} & use width as the vertical edge bound \\
\quad\raggedright\texttt{last-\allowbreak{}row-\allowbreak{}reduction-\allowbreak{}skip} & skip only the final item of a simple reduction loop \\
\quad\raggedright\texttt{last-\allowbreak{}row-\allowbreak{}reduction-\allowbreak{}skip-\allowbreak{}start1} & omit the last item while retaining a separately seeded first item \\
\quad\raggedright\texttt{le2lt} & $\le$ $\to$ $<$ : off-by-one underrun \\
\quad\raggedright\texttt{line-\allowbreak{}count-\allowbreak{}one-\allowbreak{}extra-\allowbreak{}lane} & Mechanism 3 (shape-alignment evasion): aligned line counts launch no extra lane; a non-multiple line count with guarded allocation kills it. \\
\quad\raggedright\texttt{literal-\allowbreak{}loop-\allowbreak{}last-\allowbreak{}skip} & omit the last tap of a fixed-size pooling or convolution loop \\
\quad\raggedright\texttt{loop-\allowbreak{}bound-\allowbreak{}minus1} & drop last loop iteration (last tile / last row missing) \\
\quad\raggedright\texttt{loop-\allowbreak{}start-\allowbreak{}1} & skip first loop iteration \\
\quad\raggedright\texttt{lt2le} & guard/loop bound $<$ $\to$ $\le$ : off-by-one overrun \\
\quad\raggedright\texttt{matmul-\allowbreak{}a-\allowbreak{}last-\allowbreak{}k-\allowbreak{}drop} & Mechanism 1 (tolerance absorption): one of hundreds of positive dot-product terms is lost ($<$1\%); a last-K spike kills it. \\
\quad\raggedright\texttt{matmul-\allowbreak{}a-\allowbreak{}transposed-\allowbreak{}last-\allowbreak{}k-\allowbreak{}drop} & Mechanism 1 (tolerance absorption): only the last term of a long dot product is removed; a dominant last A term kills it. \\
\quad\raggedright\texttt{matmul-\allowbreak{}b-\allowbreak{}last-\allowbreak{}k-\allowbreak{}drop} & Mechanism 1 (tolerance absorption): one reduction term among hundreds is zeroed; a last-K spike in B kills it. \\
\quad\raggedright\texttt{matrix-\allowbreak{}col-\allowbreak{}bound-\allowbreak{}use-\allowbreak{}m} & guard matrix columns with the row dimension \\
\quad\raggedright\texttt{matrix-\allowbreak{}row-\allowbreak{}bound-\allowbreak{}use-\allowbreak{}n} & guard matrix rows with the column dimension \\
\quad\raggedright\texttt{min-\allowbreak{}reduction-\allowbreak{}last-\allowbreak{}row-\allowbreak{}drop} & Mechanism 2 (input-domain evasion): a random final row is rarely the minimum; placing the unique minimum there kills it. \\
\quad\raggedright\texttt{output-\allowbreak{}elements-\allowbreak{}one-\allowbreak{}extra-\allowbreak{}lane} & Mechanism 3 (shape-alignment evasion): an exactly aligned launch has no lane at output\_elements; a misaligned output plus a CUDA memory checker kills it. \\
\quad\raggedright\texttt{pool2d-\allowbreak{}last-\allowbreak{}tap-\allowbreak{}drop} & Mechanism 1 (tolerance absorption): one of 121 comparable positive pooling terms is lost (\textasciitilde{}0.83\%); a large value in the omitted tap kills it. \\
\quad\raggedright\texttt{reduce-\allowbreak{}seed-\allowbreak{}first-\allowbreak{}element} & max-reduction seeded with 0 instead of -inf: wrong only when every value in the row is negative \\
\quad\raggedright\texttt{reduce-\allowbreak{}seed-\allowbreak{}first-\allowbreak{}element-\allowbreak{}min} & min-reduction seeded with 0: wrong only for all-positive rows \\
\quad\raggedright\texttt{reduce-\allowbreak{}shift2} & tree reduction skips levels \\
\quad\raggedright\texttt{return-\allowbreak{}guard-\allowbreak{}drop} & remove early-return bounds guard (silent OOB) \\
\quad\raggedright\texttt{strided-\allowbreak{}tail-\allowbreak{}drop} & drop the final strided chunk of a reduction or output pass \\
\quad\raggedright\texttt{tail-\allowbreak{}guard-\allowbreak{}tighten} & reject exactly the final valid scalar in a guarded tail \\
\quad\raggedright\texttt{tile-\allowbreak{}last-\allowbreak{}iteration-\allowbreak{}skip} & omit only the final matrix tile \\
\quad\raggedright\texttt{width-\allowbreak{}bound-\allowbreak{}use-\allowbreak{}height} & use height as the horizontal edge bound \\
\addlinespace
\multicolumn{2}{@{}l}{\emph{Synchronization} (13 rules)} \\
\quad\raggedright\texttt{batchnorm-\allowbreak{}local-\allowbreak{}stats-\allowbreak{}warp-\allowbreak{}barrier} & Mechanism 4 (benign race): uniform random lane statistics are close; per-lane distributions with very different means kill it. \\
\quad\raggedright\texttt{batchnorm-\allowbreak{}merge-\allowbreak{}warp-\allowbreak{}barrier} & Mechanism 4 (benign race): similar Welford partials hide stale cross-warp merges; alternating lane chunks kill it. \\
\quad\raggedright\texttt{batchnorm-\allowbreak{}output-\allowbreak{}warp-\allowbreak{}barrier} & Mechanism 4 (benign race): statistics are already reduced before this mostly redundant barrier; forced warp skew kills it. \\
\quad\raggedright\texttt{groupnorm-\allowbreak{}final-\allowbreak{}output-\allowbreak{}warp-\allowbreak{}barrier} & Mechanism 4 (benign race): the preceding reduction barrier normally makes this nearly redundant; adversarial scheduling with divergent lanes kills it. \\
\quad\raggedright\texttt{groupnorm-\allowbreak{}sqdev-\allowbreak{}store-\allowbreak{}warp-\allowbreak{}barrier} & Mechanism 4 (benign race): uniform inputs give similar squared-deviation partials; a lane-localized outlier kills it. \\
\quad\raggedright\texttt{groupnorm-\allowbreak{}sum-\allowbreak{}store-\allowbreak{}warp-\allowbreak{}barrier} & Mechanism 4 (benign race): similar lane partials mask cross-warp publication races; lane-chunk contrast kills it. \\
\quad\raggedright\texttt{matmul-\allowbreak{}after-\allowbreak{}compute-\allowbreak{}warp-\allowbreak{}barrier} & Mechanism 4 (benign race): stale and next-tile random values are statistically similar; a tile-alternating high/low input kills it. \\
\quad\raggedright\texttt{matmul-\allowbreak{}after-\allowbreak{}load-\allowbreak{}warp-\allowbreak{}barrier} & Mechanism 4 (benign race): warp-local progress often sees similar positive tiles; lane/warp-skewed tile values kill it. \\
\quad\raggedright\texttt{reduction-\allowbreak{}barrier-\allowbreak{}move-\allowbreak{}before-\allowbreak{}store} & move the barrier before the shared-memory publication \\
\quad\raggedright\texttt{shared-\allowbreak{}denom-\allowbreak{}use-\allowbreak{}local} & consume the per-thread value instead of the reduced shared value \\
\quad\raggedright\texttt{shared-\allowbreak{}sqrt-\allowbreak{}use-\allowbreak{}local} & read the local accumulator in place of the synchronized reduction \\
\quad\raggedright\texttt{sync-\allowbreak{}remove} & remove barrier: shared-memory race \\
\quad\raggedright\texttt{welford-\allowbreak{}count-\allowbreak{}merge-\allowbreak{}early} & publish a merged count to the partner lane instead of the consumer \\
\addlinespace
\multicolumn{2}{@{}l}{\emph{Precision} (30 rules)} \\
\quad\raggedright\texttt{acc-\allowbreak{}fp16} & fp32 accumulator declared fp16 (paired by fixup below) \\
\quad\raggedright\texttt{avgpool121-\allowbreak{}denom-\allowbreak{}perturb} & Mechanism 1 (tolerance absorption): the relative scale error is below 0.01\%; a much tighter tolerance kills it. \\
\quad\raggedright\texttt{batchnorm-\allowbreak{}epsilon-\allowbreak{}small-\allowbreak{}bias} & Mechanism 1 (tolerance absorption): variance dominates a 1\% epsilon perturbation; sub-epsilon variance kills it. \\
\quad\raggedright\texttt{clamp-\allowbreak{}bound-\allowbreak{}perturb} & saturation bound off by 1e-3: only saturated inputs differ \\
\quad\raggedright\texttt{erf-\allowbreak{}to-\allowbreak{}tanh-\allowbreak{}approx} & exact erf GELU replaced by the tanh approximation (\textasciitilde{}1e-3 relative) \\
\quad\raggedright\texttt{expf-\allowbreak{}fast} & expf $\to$ fast-math \_\_expf approximation \\
\quad\raggedright\texttt{fdividef-\allowbreak{}fast} & true division $\to$ fast approximate division \\
\quad\raggedright\texttt{gelu-\allowbreak{}sqrt2pi-\allowbreak{}perturb} & sqrt(2/pi) constant mistyped: relative error \textasciitilde{}1e-3 \\
\quad\raggedright\texttt{gelu-\allowbreak{}tanh-\allowbreak{}const-\allowbreak{}perturb} & GELU tanh-approximation cubic coefficient off in the 4th digit \\
\quad\raggedright\texttt{groupnorm-\allowbreak{}epsilon-\allowbreak{}halved} & Mechanism 1 (tolerance absorption): ordinary random variance dwarfs epsilon; nearly constant inputs with variance below 1e-5 kill it. \\
\quad\raggedright\texttt{hardsigmoid-\allowbreak{}half-\allowbreak{}ulpish-\allowbreak{}perturb} & Mechanism 1 (tolerance absorption): a 1e-4 offset is below atol; a sub-1e-4 tolerance kills it. \\
\quad\raggedright\texttt{hardsigmoid-\allowbreak{}six-\allowbreak{}denom-\allowbreak{}perturb} & Mechanism 1 (tolerance absorption): the slope error is \textasciitilde{}0.017\%; a tight oracle near the linear-region edge kills it. \\
\quad\raggedright\texttt{hardsigmoid-\allowbreak{}slope-\allowbreak{}perturb} & hardsigmoid offset off by 5e-4: hidden by atol at small outputs \\
\quad\raggedright\texttt{invsqrt2-\allowbreak{}perturb} & 1/sqrt(2) mistyped in the erf-based GELU \\
\quad\raggedright\texttt{kahan-\allowbreak{}compensation-\allowbreak{}drop} & remove a simple Kahan compensation subtraction \\
\quad\raggedright\texttt{literal-\allowbreak{}epsilon-\allowbreak{}remove} & drop a literal small stabilizer from a simple expression \\
\quad\raggedright\texttt{matmul-\allowbreak{}acc-\allowbreak{}store-\allowbreak{}fma-\allowbreak{}nudge} & Mechanism 1 (tolerance absorption): a 0.01\% store-scale error is hidden by rtol; a tighter relative oracle kills it. \\
\quad\raggedright\texttt{rsqrt-\allowbreak{}epsilon-\allowbreak{}remove} & remove the variance epsilon safeguard \\
\quad\raggedright\texttt{selu-\allowbreak{}alpha-\allowbreak{}perturb} & SELU alpha constant digit transposition \\
\quad\raggedright\texttt{selu-\allowbreak{}alpha-\allowbreak{}small-\allowbreak{}perturb} & Mechanism 2 (input-domain evasion): the alpha is unused for positive originals; negative SELU inputs kill it. \\
\quad\raggedright\texttt{selu-\allowbreak{}scale-\allowbreak{}perturb} & SELU scale constant digit transposition \\
\quad\raggedright\texttt{selu-\allowbreak{}scale-\allowbreak{}small-\allowbreak{}perturb} & Mechanism 1 (tolerance absorption): the global relative error is under 0.01\%; a tighter relative tolerance kills it. \\
\quad\raggedright\texttt{softmax-\allowbreak{}max-\allowbreak{}subtraction-\allowbreak{}fast-\allowbreak{}remove} & remove stabilization from a fast exponential \\
\quad\raggedright\texttt{softmax-\allowbreak{}max-\allowbreak{}subtraction-\allowbreak{}remove} & remove max subtraction, causing overflow only at large magnitude \\
\quad\raggedright\texttt{softplus-\allowbreak{}cutoff-\allowbreak{}shift} & softplus linear-regime threshold moved: wrong inside [8,20] \\
\quad\raggedright\texttt{softplus-\allowbreak{}cutoff-\allowbreak{}tiny-\allowbreak{}shift} & Mechanism 2 (input-domain evasion): original values never approach the large branch cutoff; values in [19.99,20] kill it. \\
\quad\raggedright\texttt{sqrt-\allowbreak{}epsilon-\allowbreak{}remove} & remove epsilon under a square root \\
\quad\raggedright\texttt{store-\allowbreak{}fp16} & output store through fp16 roundtrip \\
\quad\raggedright\texttt{variance-\allowbreak{}unbiased-\allowbreak{}count} & use sample rather than population variance \\
\quad\raggedright\texttt{welford-\allowbreak{}second-\allowbreak{}delta-\allowbreak{}reuse} & replace Welford's corrected second delta with the first delta \\
\addlinespace
\end{longtable}

\normalsize

\section{Knowledge ladder: raw counts and prompts}
\label{app:ladder}
Table~\ref{tab:ladder-raw} gives the counts behind the rates of
Table~\ref{tab:ladder}. All rungs target the same 1{,}154 official-input
survivors across 30 problems. R1 and R3 generate one suite file per
problem per rung; R3 is scored on 1{,}092 survivors because the evaluator
skips a problem whose mutants crashed the CUDA context on two attempts (its
crash re-entry guard), which excludes 62 survivors. Generation settings are
those of the LLM-usage statement.
\begin{table}[h]
\centering
\caption{Knowledge ladder, raw counts: survivors killed over survivors
evaluated at each rung.}
\label{tab:ladder-raw}
\vspace{5pt}
\begin{tabular}{@{}lrrr@{}}
\toprule
\textbf{Rung} & \textbf{Killed} & \textbf{Evaluated} & \textbf{Rate} \\
\midrule
R0\; random fuzz & 249 & 1,154 & 21.6\% \\
R1\; naive prompt & 495 & 1,154 & 42.9\% \\
R2\; $+$ taxonomy \& ceiling & 704 & 1,154 & 61.0\% \\
R3\; $+$ white-box diffs & 626 & 1,092 & 57.3\% \\
\bottomrule
\end{tabular}
\end{table}

The briefs handed to the generator at R1 and R3 follow verbatim (the R2
rung reuses the targeted-suite generation of \S\ref{sec:synthesis}, whose
brief ships with the artifact); the invocation prepends only the assigned
problem name and the path of its substrate.

\paragraph{R1 brief.}
{\scriptsize\verbatiminput{artifact/ladder/BRIEF_R1.md}}

\paragraph{R3 brief.}
{\scriptsize\verbatiminput{artifact/ladder/BRIEF_R3.md}}

\end{document}